\documentclass[letterpaper]{article} 
\usepackage{aaai2026}  
\usepackage{times}  
\usepackage{helvet}  
\usepackage{courier}  
\usepackage[hyphens]{url}  
\usepackage{graphicx} 
\usepackage{amsmath}
\usepackage{booktabs}
\usepackage{multirow}
\usepackage {graphicx}
\usepackage{siunitx}
\usepackage{nicefrac}
\usepackage{amsfonts}
\usepackage{natbib}  
\usepackage{caption} 
\usepackage{algorithm}
\usepackage{algorithmic}

\usepackage{newfloat}
\usepackage{listings}
\DeclareCaptionStyle{ruled}{labelfont=normalfont,labelsep=colon,strut=off} 
\floatstyle{ruled}
\newfloat{listing}{tb}{lst}{}
\floatname{listing}{Listing}
\title{VaccineRAG: Boosting Multimodal Large Language Models' Immunity to Harmful RAG Samples}
\author{
    Qixin Sun\textsuperscript{\rm 1}\thanks{Equal Contribution},
    Ziqin Wang\textsuperscript{\rm 1,\rm 2}\footnotemark[1],
    Hengyuan Zhao\textsuperscript{\rm 1}\footnotemark[1],
    Yilin Li,
    Kaiyou Song,
    Linjiang Huang\textsuperscript{\rm 1},
    Xiaolin Hu\textsuperscript{\rm 2}\thanks{Corresponding author},
    Qingpei Guo\footnotemark[2],
    Si Liu\textsuperscript{\rm 1}\footnotemark[2]
}
\affiliations{
    \textsuperscript{\rm 1}School of Artificial Intelligence, Beihang University\\
    \textsuperscript{\rm 2}QIYUAN LAB\\
}

\usepackage{bibentry}

\begin{document}

\maketitle

\begin{abstract}

Retrieval Augmented Generation enhances the response accuracy of Large Language Models (LLMs) by integrating retrieval and generation modules with external knowledge, demonstrating particular strength in real-time queries and Visual Question Answering tasks. 
However, the effectiveness of RAG is frequently hindered by the precision of the retriever: many retrieved samples fed into the generation phase are irrelevant or misleading, posing a critical bottleneck to LLMs’ performance.
To address this challenge, we introduce \textbf{VaccineRAG}, a novel Chain-of-Thought-based retrieval-augmented generation dataset. 
On one hand, VaccineRAG employs a benchmark to evaluate models using data with varying positive/negative sample ratios, systematically exposing inherent weaknesses in current LLMs. 
On the other hand, it enhances models’ sample-discrimination capabilities by prompting LLMs to generate explicit Chain-of-Thought (CoT) analysis for each sample before producing final answers.
Furthermore, to enhance the model’s ability to learn long-sequence complex CoT content, we propose \textbf{Partial-GRPO}. 
By modeling the outputs of LLMs as multiple components rather than a single whole, our model can make more informed preference selections for complex sequences, thereby enhancing its capacity to learn complex CoT.
Comprehensive evaluations and ablation studies on VaccineRAG validate the effectiveness of the proposed scheme. The code and dataset will be publicly released soon.

\end{abstract}

\section{Introduction}
In the era of rapid information growth, maintaining the currency of large-scale models is a formidable challenge. 
The overwhelming and ever-increasing amount of online data necessitates continuous model updates to ensure their outputs remain accurate and relevant.
Traditional model finetuning ~\cite{hu2022lora,liu-etal-2022-p,li2021prefix,houlsby2019parameter,karimi2021compacter,chen2023octavius} is resource-heavy and time-consuming, making frequent refreshes impractical.
Retrieval-Augmented Generation (RAG)~\cite{NEURIPS2020_6b493230,lewis2020retrieval,gao2023retrieval} tackles this challenge by retrieving query-relevant external knowledge from online or offline databases and feeding these results into LLMs for response generation.
However, due to the uncertainty of external knowledge, a critical issue is posed in current RAG systems: their retrieval mechanisms often prioritize speed over accuracy when handling vast knowledge bases~\cite{chen-etal-2024-m3}. 
Consequently, the subsequent generation is easily misled by spurious evidence, especially content that is lexically similar but semantically incongruent, ultimately compromising the accuracy of generated responses.

To address this issue, prior works~\cite{DBLP:journals/corr/abs-2407-21439, lin2025mmembed,he2024g, jiang2024longrag} have focused on improving the retriever or designing specialized retrieval pipelines, to enhance retrieval quality and reduce the number of harmful samples entering the generation phase. 
MM-Embed~\cite{lin2025mmembed} proposes modality-aware hard negative mining to mitigate the modality bias exhibited by MLLM retrievers, and  RagVL~\cite{DBLP:journals/corr/abs-2407-21439} proposes a multimodal RAG framework using MLLM as a Re-ranker to pick positive retrieval samples.
Although these retrieval-centric methods can reduce the incidence of harmful evidence, they implicitly assume that the retriever’s precision remains stable after deployment. 
In practice, retrieval quality often degrades due to various factors, such as domain shifts caused by continuous updates to the knowledge base. 
Therefore, LLMs remain vulnerable to spurious passages, and their downstream performance suffers accordingly.

SURf introduces a self-refinement framework to enhance the robustness against irrelevant retrieval samples without improving the retriever~\cite{sun-etal-2024-surf}. 
By jointly providing the LLMs both relevant and irrelevant retrieved samples and supervising it with the ground-truth answer, this approach suppresses the influence of harmful samples.
However, it still suffers from two key drawbacks:
\textbf{1) Lack of diagnostic signal}, which makes error attribution difficult and impedes the fine-grained optimization of the model's evidence-selection behavior. 
\textbf{2) Slow evidence alignment}, when dozens of retrieved samples are included, the model is forced to learn from a single scalar loss derived from the ground-truth answer. It may result in sparse gradients across all tokens, leading to slow and data-inefficient training.

\begin{figure*}[h]
  \centering
  \includegraphics[width=0.9\textwidth]{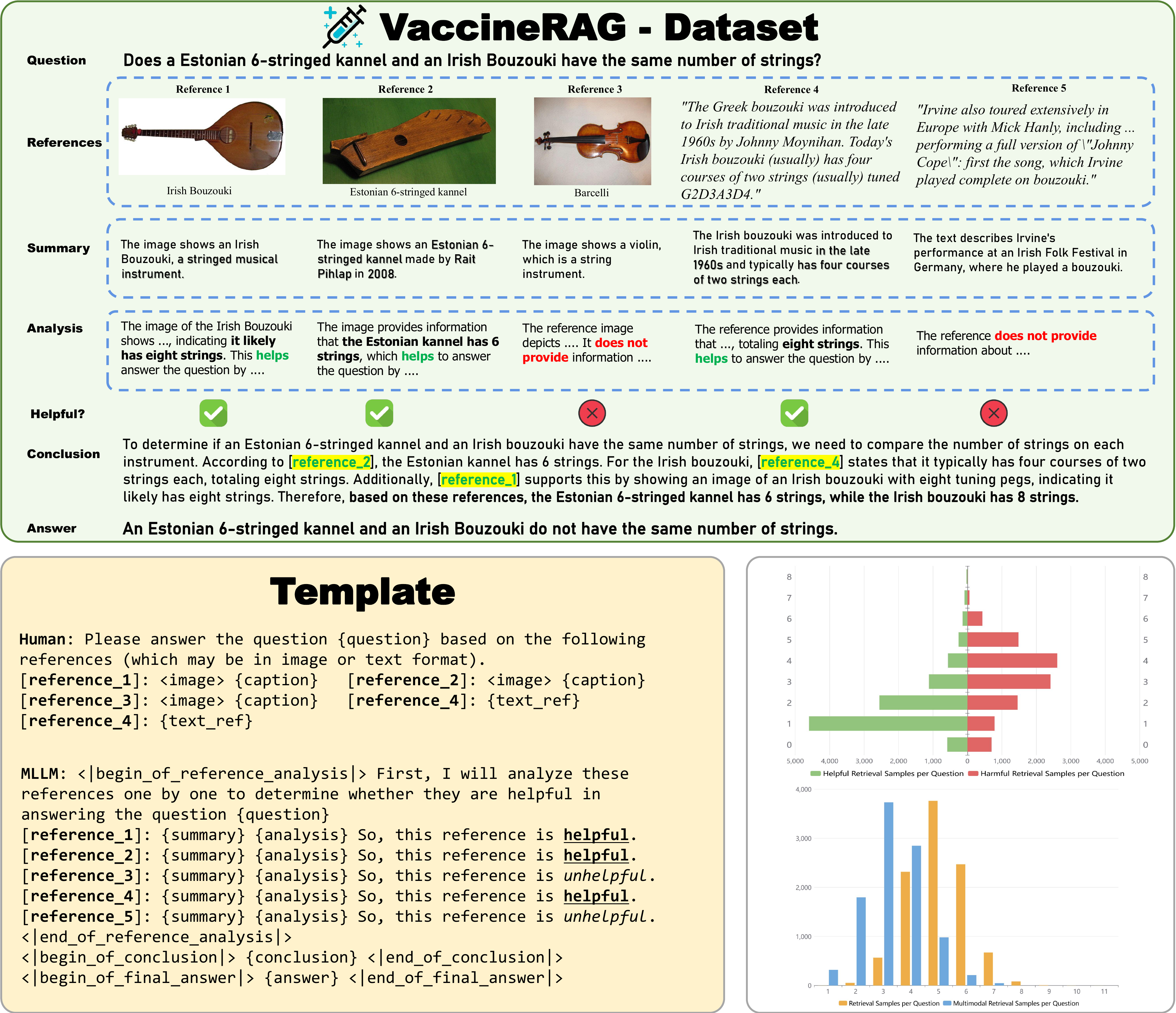}
  \caption{Top: Our multimodal RAG dataset VaccineRAG annotated with CoT. The reasoning process encompasses summarizing, analyzing each retrieved sample in conjunction with the original query, integrating the helpful samples, and ultimately generating the final answer. Left Down: Template we used for training; Right Down: Preliminary statistics of the dataset.}
  \label{img:dataset}
\end{figure*}

In this paper, we pioneer leveraging the deep reasoning capabilities of LLMs through their Chain-of-Thought reasoning scheme to alleviate the aforementioned issue.
Given a series of retrieval results containing positive and negative samples, LLMs are tasked with diagnosing the helpfulness of each retrieved sample, summarizing relevant ones, and generating the final answer step by step.
To furnish dense supervision and thus richer gradient signals throughout the reasoning process, we introduce the first CoT-based Multi-Modal RAG dataset as shown in Figure~\ref{img:dataset}, named \textbf{VaccineRAG}.
Each instance pairs an initial question with its corresponding image and descriptive caption, alongside a \emph{helpfulness} label, collectively prompting the model to generate a structured, multi-step chain-of-thought analysis.
Consequently, we have constructed a dataset comprising 10k entries, with each entry containing an average of five retrieved samples.

\begin{figure*}[t]
\centering
\includegraphics[width=\textwidth]{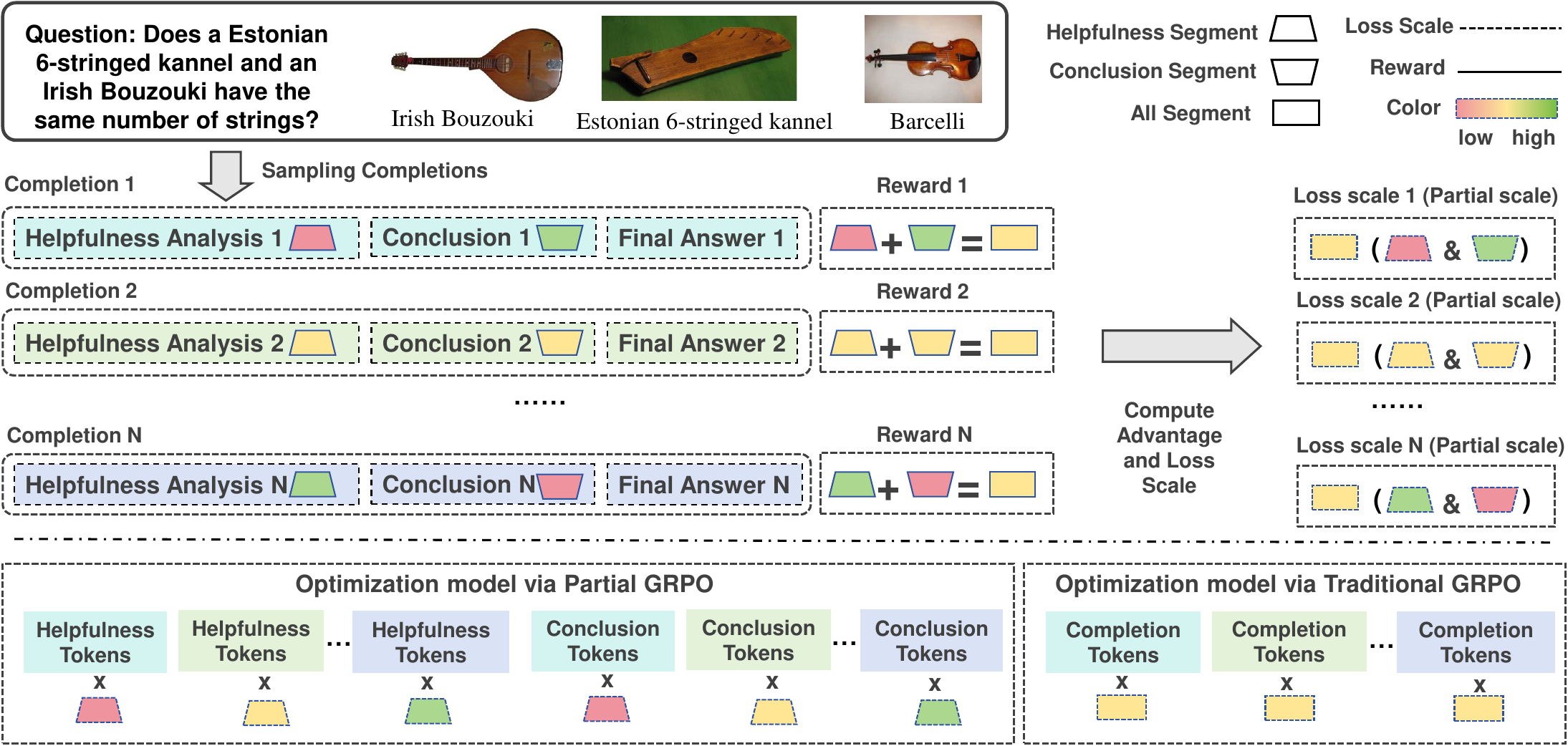}
  \caption{Partial GRPO calculates the objective function by multiplying the rewards applied to different parts of the completion with the importance sampling ratio of the related tokens, thus providing more fine-grained reward signals. The figure shows that traditional GRPO optimization cannot distinguish a better completion, whereas partial GRPO can identify sections that are not globally but locally improved and encourage the model to generate them. The Format Reward is omitted in the figure because its calculation method is the same as in traditional GRPO.}
  \label{pgrpo}
\end{figure*}

We employ the training paradigms widely adopted in contemporary MLLMs, specifically SFT followed by RL, e.g. GRPO~\cite{shao2024deepseekmath} for network optimization~\cite{chen2024internvl,bai2025qwen25vltechnicalreport}. 
However, the vanilla GRPO treats rewards accrued by all tokens in the sequence uniformly.
As illustrated in Figure~\ref{pgrpo}, when conducting separate preference optimization for distinct segments of the CoT, vanilla GRPO is prone to advantage function misalignment: for instance, low-reward segments may reduce the loss magnitude of tokens in high-reward segments, thereby leading to slow convergence and performance degradation.
To address this, we propose \textbf{Partial-GRPO}, in which the optimization process enables targeted gradient backpropagation tailored to tokens from distinct segments of the CoT, thereby significantly accelerating the convergence speed of post-optimization and improving the final model performance.

During the experimental phase, we selected three mainstream multimodal large models as the base models. We optimized them on our proposed VaccineRAG dataset using Partial GRPO, and compared the results with several constructed baselines (including SURf). The comparison verified that our proposed method can effectively counter harmful retrieved samples and improve generative robustness in multimodal RAG tasks. Furthermore, ablation studies on the components of our method demonstrated the usefulness of each structure. Examples of using our method for RAG can be found in the appendix.

\section{Related work}

\noindent\textbf{Multimodal Retrieval-Augmented Generation.}

Retrieval-augmented generation (RAG) was initially proposed to tackle knowledge-intensive tasks in natural language processing (NLP)~\cite{NEURIPS2020_6b493230,lewis2020retrieval,gao2023retrieval}. By integrating knowledge retrieved from external sources during generation, RAG has achieved notable success and widespread use in NLP~\cite{asai2024selfrag,ram-etal-2023-context,gao2024retrievalaugmentedgenerationlargelanguage,lan2023copy,chen-etal-2024-dense}. Recently, RAG has also attracted growing attention in multimodal large models~\cite{shohan-etal-2024-xl,yan-xie-2024-echosight,zhao-etal-2024-unifashion}. For example, in Visual Question Answering, Wiki-LLaVA~\cite{Caffagni_2024_CVPR} employs a hierarchical RAG mechanism to retrieve relevant Wikipedia passages, enhancing the model's ability to answer complex, domain-specific questions. In Captioning, RA-TX~\cite{10.1145/3663667} leverages captions from similar images to generate more accurate and context-aware descriptions. However, recent studies~\cite{hu2025mragbench,sun-etal-2024-surf} have shown that low-quality retrieved samples may introduce noise and degrade generation quality, highlighting the need for methods that improve model robustness against such detrimental retrievals.

\noindent\textbf{Chain of Thought (CoT) and Reinforcement Learning.}

The CoT approach enhances the reasoning capability of LLMs by prompting them to sequentially derive intermediate steps rather than directly producing the final output, thereby improving their performance on complex problem-solving tasks~\cite{NEURIPS2022_9d560961}. This enhancement significantly improves the reasoning capabilities, interpretability, controllability, and flexibility of large-scale models~\cite{yu2023betterchainofthoughtpromptingstrategies,qiao-etal-2023-reasoning,lu-etal-2023-survey}. In the domain of multimodal large models, the CoT-based approach is frequently employed to address complex, multi-step problems across various domains~\cite{zhang2023multicot,xu2024llavacot,NEURIPS2022_11332b6b}. The latest Deepseek-R1~\cite{shao2024deepseekmath} integrates reinforcement learning with the CoT approach, endowing the model with the capability of self-sampling and evolution. For the problem investigated in this study, which also involves reasoning in complex knowledge and tasks to determine the relationship between retrieved samples and the original question, we can consider employing reinforcement learning to enable the large model to perform step-by-step reasoning and achieve the desired outcomes.
\section{Preliminary}

A RAG system typically consists of two components: a retriever, responsible for retrieving samples relevant to the input from a database, and a generator, which takes the retrieved samples and the original input concatenated together to perform the generation. 
For an input $x$ and a database $\mathcal{D} = \{d_i\}_{i=1}^{|\mathcal{D}|}$, the retriever typically employs maximum inner product search to match the top $\text{K}$ samples from the database that have the highest similarity to the input embedding. 
The final set of retrieved samples is given by:
\begin{equation}
    \mathcal{S}_{\text{retrieval}} = \mathop{\text{argTopK}}_{d \in \mathcal{D}} \left( \text{Emb}(x) \cdot \text{Emb}(d) \right),
\end{equation}
where $\text{Emb}(\cdot)$ is the embedding function. 
The output $y_{t+1}$ of the LLM (denoted as $\pi$) at time step $t+1$ is:
\begin{equation}
    y_{t+1} = \pi\left( x; \mathcal{S}_{\text{retrieval}}; y_t \right).
\end{equation}

However, given the large database size and the need to compute against input $x$ for each sample, retrievers are typically designed to be very simple to ensure acceptable retrieval times, resulting in retrieved samples of inconsistent quality.
Consequently, irrelevant samples or those lacking useful information may interfere with the model, leading to incorrect answers.

\section{VaccineRAG}

\subsection{Data Construction}

To teach MLLMs to analyze and utilize retrieval samples for multimodal RAG using the CoT approach, we leverage the WebQA dataset~\cite{chang2022webqa} as the basis for constructing a new dataset. 
Each sample in this dataset includes a text-based question, multiple retrieval results, and a label indicating their helpfulness. However, the dataset does not provide an explanation of the helpfulness of each retrieval sample, nor does it indicate how the final answer is derived from these retrieval samples. In fact, these two steps correspond to two levels of reasoning:

1) \textbf{Reasoning within retrieval samples}: This involves summarizing the content of a retrieval sample and analyzing its relevance to the original question in order to determine whether the sample is helpful.

2) \textbf{Cross-retrieval sample reasoning}: This involves integrating the helpful retrieval samples from the previous stage and using them to infer the answer to the original question.

To obtain these CoT annotations, we use SOTA commercial large model for annotation, followed by manual verification. Specifically, for each sample, we use GPT-4o to perform the following annotations: 1) Retrieval Sample Summarization: For all retrieval results in a sample, we input images (if available) to obtain their descriptions. 2) Helpfulness Analysis: Using the sample's helpfulness label and the original question as prompts, we derive an analysis to determine whether the sample is genuine. 3) Final Conclusion: After synthesizing the analysis of all retrieval results, we generate a conclusion that aligns with the final answer.

\subsection{Manual Verification} We also performed manual verification of the information generated by GPT-4o to ensure annotation quality. For example, for the Helpfulness Analysis annotation, we compared the Helpfulness generated by GPT-4o with the Helpfulness annotated in the original WebQA. If they matched, we had strong confidence in the correctness of the annotation. If there was a discrepancy, manual verification was conducted, and corrections or exclusions were made accordingly. We provided a standardized template to link the above information together, forming a natural language-based CoT for model training.

\subsection{Dataset Statistic}
In total, we constructed approximately 10,000 samples. The dataset samples, the templates utilized in our study and some statistical details are illustrated in the Figure~\ref{img:dataset}. Statistically, this dataset comprises approximately 10k entries, with each entry containing an average of $5.05 \pm 1.06$ retrieved samples. Among these, the number of image-type references averages $3.32 \pm 1.08$, while the count of helpful retrieved samples is $1.85 \pm 1.36$. Notably, 84\% of the questions are answerable, whereas the remaining questions cannot be addressed due to insufficient information. For additional examples and further statistical details regarding this dataset, please refer to the supplementary materials.

\section{Methodology}

In this section, we introduce our training scheme, consisting of two successive stages: SFT-based model warming up for output knowledge alignment and Partial-GRPO-based model preference optimization.
\subsection{Stage \uppercase\expandafter{\romannumeral1}: Warming Up}
\label{section:stage1}
To enable the base model to adapt to the task and the basic format of the output, thereby preventing the generation of outputs that fail to meet the fundamental formatting requirements during the next stage, we first optimize the base model using Supervised Fine-Tuning.
To mitigate the risk of excessive training, which could otherwise lead to a loss of sampling diversity in the subsequent stage, we just utilize 15\% of the data and train within a few steps.

\subsection{Stage \uppercase\expandafter{\romannumeral2}: Preference Optimization}
\label{section:stage2}
\subsubsection{Vanilla-GRPO}

Given an input $q$, traditional GRPO samples a group of $G$ completions $\{o_i\}_{i=1}^{G}$ from the old policy $\pi_{\text{old}}$. 
Let the partial importance sampling ratio for a certain completion \( o_i \) in the interval \([x_1, x_2]\)($x_1\geq1, x_2\leq|o_i|$) be 
\begin{equation}
\begin{aligned}
\omega_{x_1}^{x_2}(o_i) = \frac{1}{x_2-x_1}\sum_{t=x_1}^{x_2} \frac{\pi_\theta(o_{i,t} | q, o_{i,<t})}{\pi_{\theta_{\text{old}}}(o_{i,t} | q, o_{i,<t})}.
\end{aligned}
\end{equation}
Then, traditional GRPO optimizes the policy model by maximizing the given objective:

\begin{equation}
\begin{aligned}
\mathcal{J}_{\text{GRPO}}(\theta)
&= \mathbb{E}_{q \sim P(Q),\, \{o_i\}_{i=1}^G \sim \pi_{\theta_{\text{old}}}(O|q)}
\\
&\quad \left[
    \frac{1}{G} \sum_{i=1}^G \omega_{1}^{|o_i|}(o_i) \hat{A}_{i}
\right],
\end{aligned}
\end{equation}
To improve readability, we omitted the KL divergence term and the clipping objective function in the formula.
Here, \(\hat{A}_{i}\) represents the relative advantage of the \(i\)-th completion, which can be calculated by normalizing the advantages across a group of completions. For more details, you can find them in the appendix.

\subsubsection{Partial-GRPO}

Directly calculating the total reward function through a weighted sum and applying it to evaluate the entire completion, although reasonable, wastes the information about the quality of different parts in the CoT. 
In our Partial-GRPO, there are mainly two core designs: 1)  Reward Function and 2) Gradient backpropagation. 

\paragraph{Reward functions.}
 1) Format Reward. The format reward \(r^f\) is designed to encourage the model to generate completions that adhere to the dataset template. A reward of 1 is assigned to completions that conform to the required format, while a reward of 0 is given otherwise. The scope of this reward function is the whole completion.
\begin{equation}
        r^f = 
        \begin{cases} 
        1, & \text{if the completion adheres to the format} \\
        0, & \text{otherwise}
        \end{cases}.
        \end{equation}

2) Helpfulness Reward. The helpfulness reward \(r^h\) is designed to evaluate whether the model correctly analyzes the helpfulness of each retrieved sample. For an input containing \(n\) retrieved samples, we extract the keywords (``helpful''/``unhelpful'') from the model's analysis of each retrieved sample to determine the model's judgment on its helpfulness. Let the ground truth of the helpfulness for the \(j\)-th retrieved sample be denoted as \(H_j^{gt}\), and the model's judgment as \(H_j\). The sub-reward function for each retrieved sample, \(r_{h,j}\), is defined as:
\begin{equation}
        r^h_{j} = 
        \begin{cases} 
        1, & \text{if}\;H_j=H_j^{gt} \\
        0, & \text{otherwise}
        \end{cases}.
        \end{equation}
Then, we can get the overall helpfulness reward function $r^h$ as below
\begin{equation}
r^h=\frac{1}{n}\sum_{j=1}^n r^h_{j}.
        \end{equation}
The scope of this reward function is the helpfulness analysis section.

 3) Conclusion Reward. The conclusion reward \(r^c\) is utilized to verify whether the model correctly cites the appropriate retrieved samples in the summary section. Specifically, it evaluates whether the model cites the samples that were previously determined as ``helpful'' in the reference analysis section, while refraining from citing those deemed ``unhelpful''.
For the \(j\)-th retrieved sample, if it is cited in the conclusion section of the completion, \(C_j\) is assigned a value of true; otherwise, it is assigned false. The sub-reward function for this retrieved sample is defined as:
\begin{equation}
        r^c_{j} = 
        \begin{cases} 
        1, & \text{if}\;C_j=H_j \\
        0, & \text{otherwise}
        \end{cases}.
        \end{equation}
Similarly, the overall conclusion reward function is defined as: 
\begin{equation}
r^c=\frac{1}{n}\sum_{j=1}^n r^c_{j}.
\end{equation}
The scope of this reward function is the conclusion section.
\paragraph{Gradient backpropagation.}
Given the i-th sampled completion \( o_i \), it is divided into three parts: helpfulness analysis, conclusion, and final answer, which are denoted as \( o_i^h \), \( o_i^c \), and \( o_i^f \), respectively.

\begin{equation}
\begin{aligned}
\mathcal{J}_{\text{P-GRPO}}(\theta)
&= \mathbb{E}_{q \sim P(Q),\, \{o_i\}_{i=1}^G \sim \pi_{\theta_{\text{old}}}(O|q)}
\\
&\quad \left[
    \frac{1}{G} \sum_{i=1}^G  \hat{\boldsymbol{R}}_i \boldsymbol{{\Omega}}_i
\right],
\end{aligned}
\end{equation}

Among them, $\hat{\boldsymbol{R}}$ and $\boldsymbol{{\Omega}}_i$ are respectively the three normalized rewards  of the $i$-th completion and the partial importance sampling ratios corresponding to the relative tokens (or ``scope'') of each reward.
\begin{equation}
\hat{\boldsymbol{R}}_i=
\begin{bmatrix} 
\hat{r}^{h}(o_i)\\ \hat{r}^{c}(o_i) \\ \hat{r}^{f}(o_i)
\end{bmatrix}^\top , 
\boldsymbol{{\Omega}}_i=
\begin{bmatrix} 
\omega_{1}^{|o_i^h|}(o_i) \\\omega_{|o_i^h|+1}^{|o_i^h|+|o_i^c|}(o_i) \\
\omega_{1}^{|o_i|}(o_i)
\end{bmatrix},
\end{equation}

where $\hat{r}^{h}(o_i)$ is the helpfulness reward of the $i$-th completion after normalization across a group of completions, and $\hat{r}^{c}(o_i)$ and $\hat{r}^{f}(o_i)$ are defined similarly.

\section{Experiments}
\begin{table*}[h]
\centering
\caption{Performance of our approach and baselines on polluted generation}
\renewcommand\arraystretch{1}
\renewcommand\tabcolsep{4.0pt}
\label{tab:acc_adr5_ours}
  \begin{tabular}{ll|cccccc|cc}
    \toprule
    \textbf{Model} & \textbf{Type} 
      & $\text{ACC}_{+0}$ & $\text{ACC}_{+1}$ & $\text{ACC}_{+2}$
      & $\text{ACC}_{+3}$ & $\text{ACC}_{+4}$ & $\text{ACC}_{+5}$ & $\text{MA}_{5}\uparrow$ & $\text{ADR}_5\downarrow$ \\
\midrule

\multirow{5}{*}{Qwen2-VL} & Zero-shot & 46.81 & 44.15 & 41.03 & 39.53 & 37.93 & 37.34 & 41.13 & 34.05 \\ 
 & SURf       & 49.46 & 47.39 & 44.48 & 44.66 & 43.59 & 42.12 & 45.28 & 25.08 \\ 
 & SURf+CoT     & 50.35 & 46.92 & 47.44 & 48.04 & 47.34 & 45.41 & 47.58 & 16.60 \\ 
 & GRPO      & 58.46 & 54.94 & 54.26 & 53.72 & 53.03 & 52.60 & 54.50 & 23.75 \\ 
  & Partial GRPO      & \textbf{58.93} & \textbf{56.24} & \textbf{55.64} & \textbf{55.72} & \textbf{55.50} & \textbf{55.58} & \textbf{56.27} & \textbf{15.98} \\ 
\midrule
\multirow{5}{*}{Qwen2.5-VL} & Zero-shot & 62.42 & 55.50 & 55.52 & 54.66 & 52.19 & 51.79 & 55.35 & 42.45 \\ 
 & SURf       & 61.89 & 56.32 & 56.22 & 57.16 & 56.06 & 54.69 & 57.06 & 29.00 \\ 
 & SURf+CoT     & 60.82 & 59.44 & 60.19 & 61.12 & 59.49 & 58.06 & 59.85 & 5.83 \\ 
 & GRPO      & 59.86 & 59.44 & 60.19 & 61.11 & 59.49 & 58.86 & 59.83 & \textbf{0.21} \\ 
  & Partial GRPO      & \textbf{66.27} & \textbf{64.97} & \textbf{64.02} & \textbf{64.28} & \textbf{63.92} & \textbf{63.73} & \textbf{64.53} & 10.43 \\ 
\midrule
 \multirow{5}{*}{InternVL3} & Zero-shot & 62.31 & 58.37 & 57.27 & 57.09 & 56.18 & 54.99 & 57.70 & 27.65 \\ 
 & SURf       & 49.02 & 42.28 & 45.64 & 41.28 & 42.75 & 44.48 & 44.24 & 28.67 \\ 
 & SURf+CoT     & 57.49 & 55.13 & 55.02 & 54.85 & 54.64 & 54.37 & 55.25 & 13.44 \\ 
& GRPO      & 63.47 & 62.02 & 60.43 & 60.06 & 58.67 & 59.02 & 60.61 & 17.13 \\ 
  & Partial GRPO      & \textbf{64.24} &\textbf{ 63.18} & \textbf{62.60} & \textbf{61.74} & \textbf{61.49} & \textbf{60.43} & \textbf{62.28} & \textbf{11.76} \\ 
\bottomrule
\end{tabular}
\end{table*}
\subsection{Setups and Implementation Details.}
\textbf{Experiments Setup.} To evaluate the MLLM's immunity against harmful retrieval samples, we selected approximately 1,000 questions from the validation set of WebQA. The 1,000 questions were divided into two distinct groups: answerable (i.e., those for which correct answers can be derived from the helpful retrieval samples, 858 questions) and unanswerable (i.e., those for which the information provided by the helpful retrieval samples is insufficient to formulate an answer). We designed two strategies for testing on answerable questions: 1) \textbf{Polluted generation}: For each answerable question, harmful retrieval samples are incrementally added to the helpful retrieval samples.  This strategy assesses the impact of additional harmful samples on the generation when helpful samples are retrieved. 2) \textbf{TopK generation}: Using the retriever, a fixed number of $K$ samples are retrieved from the database for each answerable question to facilitate generation. This strategy directly evaluates the final effectiveness of the RAG framework.
For experiments related to unanswerable questions, please refer to the supplementary materials.

\vspace{1mm}
\noindent \textbf{Implementation Details.}
We used $8\times96\text{GB}$ NVIDIA H20 GPUs for both training and inference. For additional details  such as learning rates, batch sizes, and other hyperparameters, please refer to the supplementary materials.

\subsection{Metrics}

\textbf{Accuracy ($\text{ACC}$).} Given that the employed VQA dataset is open-ended in nature, we utilized the closed-source model GPT-4o to evaluate the accuracy of the answers. Specifically, based on the triad of question, ground truth, and the model's generated answer, each response was scored span from 0 to 5. Subsequently, these scores were normalized to a range of 0 to 100 for comprehensive assessment. 

Through experiments, we found that using this method for evaluation aligns much more closely with human assessments compared to the method used in WebQA. This is because the questions in WebQA are open-ended, allowing for multiple possible answers, which leads to misjudgments in the native evaluation method used in WebQA. Moreover, for the same input, the scores obtained using this evaluation method are consistent across multiple assessments. Therefore, we believe this scoring approach is reasonable.

\vspace{1mm}
\noindent\textbf{Mean Accuracy ($\text{MA}$).} For polluted generation, when the number of harmful retrieval samples added is $k$, and the overall accuracy is $\text{ACC}_{+k}$, the average accuracy is denoted as:
\begin{equation}
    \text{MA}_k=\frac{1}{k} \sum_{i=1}^k \text{ACC}_{+i}.
\end{equation}

\vspace{1mm}
\noindent \textbf{Accuracy Degradation Rate ($\text{ADR}$).} For polluted generation, when the number of harmful retrieval samples added is $k$, and the overall accuracy is $\text{ACC}_{+k}$, the accuracy degradation rate $\text{ADR}_k$ is calculated as follows: 
\begin{equation}
    \text{ADR}_k=\frac{1}{k} \sum_{i=1}^k \frac{\text{ACC}_{+0}-\text{ACC}_{+i}}{\text{ACC}_{+0}}.
\end{equation}

\subsection{Baselines}
We selected Qwen2-VL-7B ~\cite{bai2025qwen25vltechnicalreport}, Qwen2.5-VL-7B ~\cite{wang2024qwen2vlenhancingvisionlanguagemodels}, and InternVL3-8B ~\cite{chen2024internvl} as the base models for training. The following methods were chosen as baselines to compare with our proposed approach: 

\hspace{2mm}\textbf{1) Zero-shot}: This approach involves using the base model without any fine-tuning, directly prompting it to generate the final answer.

\hspace{2mm}\textbf{2) SURf}: According to the approach in SURf~\cite{sun-etal-2024-surf}, this method does not utilize the CoT information from VaccineRAG, but instead employs only the final answer as supervision for SFT.

\hspace{2mm}\textbf{3) SURf+CoT}: Based on the SURf method, we introduce a CoT process to reason about the retrieved samples.

\hspace{2mm}\textbf{4) GRPO}: Using the traditional GRPO algorithm, multiple reward functions are weighted and summed to evaluate the entire completion segment, and this evaluation is used as the advantage in GRPO.

\subsection{Main Result}

\begin{table}[htb]
  \centering
  \caption{The accuracy of our approach and baselines, varies with the number of retrieved samples}
  \label{tab:topk_generation}

  \begin{tabular}{lcccc}
    \toprule
     \textbf{Type}  & \multicolumn{4}{c}{$\text{ACC}_{\text{Top}-k}$} \\
    \midrule
     k & 2&5&10&15 \\
    \midrule
    \multicolumn{5}{l}{\textit{\textbf{Qwen2-VL}}} \\
     Zero-shot        & 28.39 & 30.68 & 30.26 & 28.93 \\
     SURf       & 28.39 & 31.61 & 31.77 & 30.70 \\
     SURf+CoT  & 24.94 & 31.14 & 34.97 & 33.57 \\
     GRPO          & 32.58 & 36.87 & 40.62 & 40.09 \\
     Partial GRPO          & \textbf{33.74} & \textbf{38.60} & \textbf{42.27} & \textbf{41.44} \\
    \midrule
    \multicolumn{5}{l}{\textit{\textbf{Qwen2.5-VL}}} \\
     Zero-shot        & \textbf{37.60} & 40.12 & 40.84 & 39.98 \\
    SURf       & 27.76 & 33.45 & 38.07 & 40.68 \\
    SURf+CoT  & 37.90 & 39.58 & 44.04 & 42.94 \\
    GRPO          & 31.76 & \textbf{41.52} & 43.98 & 44.75 \\
     Partial GRPO          & 33.87 & 40.69 & \textbf{45.84} & \textbf{47.21} \\
     \midrule
    \multicolumn{5}{l}{\textit{\textbf{InternVL3}}} \\
     Zero-shot        & 26.50 & 31.82 & 33.40 & 37.69 \\
    SURf       & 17.16 & 19.14 & 22.56 & 23.96 \\
    SURf+CoT  & 24.57 & 30.02 & 36.83 & 38.93 \\
    GRPO          & \textbf{28.95} & 36.62 & 41.03 & 43.17 \\
     Partial GRPO          & 28.76 & \textbf{36.85} & \textbf{42.74} & \textbf{46.08} \\
    \bottomrule
  \end{tabular}

\end{table}
\begin{table*}[t]
\centering
\caption{Performance of Partial GRPO and its variants with certain rewards removed on polluted generation. The variant with the format reward removed is not shown in the table due to unstable training.}
\renewcommand\arraystretch{1}
\renewcommand\tabcolsep{4.0pt}
\label{tab:ablation_1}
  \begin{tabular}{ll|cccccc|cc}
    \toprule
    \textbf{Model} & \textbf{Type} 
      & $\text{ACC}_{+0}$ & $\text{ACC}_{+1}$ & $\text{ACC}_{+2}$
      & $\text{ACC}_{+3}$ & $\text{ACC}_{+4}$ & $\text{ACC}_{+5}$ & $\text{MA}_{5}\uparrow$ & $\text{ADR}_5\downarrow$ \\
\midrule

\multirow{3}{*}{Qwen2-VL} 
  & Partial GRPO      & \textbf{58.93} & \textbf{56.24} & \textbf{55.64} & \textbf{55.72} & \textbf{55.50} & \textbf{55.58} & \textbf{56.27} & \textbf{15.98} \\ 
  & w/o Helpfulness Reward      & 55.84 & 53.31 & 52.22 & 51.79 & 52.03 & 51.04 & 52.71 & 18.80 \\ 
  & w/o Conclusion Reward      & 56.62 & 55.31 & 54.01 & 53.12 & 53.46 & 50.88 & 53.90 & 16.34 \\ 
\midrule
\multirow{3}{*}{Qwen2.5-VL} & Partial GRPO & \textbf{66.27} &\textbf{ 64.97} & \textbf{64.02} & \textbf{64.28} & \textbf{63.92} & \textbf{63.73} & \textbf{64.53} & \textbf{10.43} \\ 
 & w/o Helpfulness Reward      & 60.44 & 58.02 & 57.95 & 58.20 & 58.60 & 57.41 & 58.44 & 12.03 \\ 
  & w/o Conclusion Reward      & 62.28 & 61.17 & 60.26 & 61.59 & 59.15 & 57.26 & 60.29 & 11.99 \\ 
\midrule
 \multirow{3}{*}{InternVL3} & Partial GRPO & \textbf{64.24} & \textbf{63.18} & \textbf{62.60} & \textbf{61.74} & \textbf{61.49} & \textbf{60.43} & \textbf{62.28} & \textbf{11.76} \\ 
 & w/o Helpfulness Reward      & 57.82 & 51.74 & 53.73 & 50.35 & 50.68 & 50.44 & 52.46 & 31.18 \\ 
  & w/o Conclusion Reward      & 60.49 & 60.26 & 58.62 & 56.95 & 58.62 & 57.13 & 58.68 & 10.87 \\  
\bottomrule
\end{tabular}
\end{table*}
\paragraph{Polluted Generation.} To evaluate polluted generation, we systematically injected 0-5 harmful retrieval samples into the set of helpful retrieval samples, measured the accuracy of the multimodal large model’s responses, and then calculated the mean accuracy and the accuracy degradation rate.

The results are shown in Table~\ref{tab:acc_adr5_ours}. Without training, all three models tested experienced varying degrees of accuracy degradation when harmful retrieval samples were added. For example, Qwen2.5-VL achieved the highest accuracy of 62.42\% without any harmful samples, but its Accuracy Degradation Rate reached 42.45\% after five harmful retrieval samples were added, the worst among the three models. On the other hand, although InternVL3 performed slightly worse than Qwen2.5-VL without any harmful samples, it exhibited the best performance in terms of Accuracy Degradation Rate. \textbf{This indicates that the performance decline caused by the inclusion of harmful samples is a common issue in the RAG tasks of tested multimodal large models on the dataset we proposed.}

After training with the Partial GRPO method we proposed, the model's Accuracy Degradation Rate showed significant improvement compared to other baselines, without any loss in Mean Accuracy. 
\textbf{This suggests that our method enhances the model's robustness when harmful retrieval samples are added.}
Additionally, we observed that while the SURf method also reduces accuracy loss when harmful retrieval samples are included to some extent, it leads to a loss in Mean Accuracy. Adding CoT helped mitigate this loss, as SURf lacks the analytical context of CoT as guidance, making it difficult to learn the relationship between the final answer and the polluted retrieval samples.

\paragraph{TopK Generation.} We employed BGE-VL-base~\cite{zhou2024megapairs} as the retriever, utilizing the questions directly as queries. From the knowledge base constructed by all references in the validation split of WebQA, we selected the Top-K (K=2,5,10,15) samples through maximum inner product search to participate in the generation process.

The experimental results are shown in Table~\ref{tab:topk_generation}. When using a larger $\text{K}$ value, such as $\text{K}=10$ or $\text{K}=15$, our trained model consistently achieved the best performance compared to other baselines, and also attained higher accuracy than with smaller K values. 
\textbf{This indicates that when provided with more retrieval samples, the model is able to effectively utilize the helpful ones while being less affected by harmful samples.}
In contrast, for untrained models like Qwen2.5-VL and Qwen2-VL, accuracy declined to some extent as the K value increased. This further confirms that our proposed method enhances the model's robustness against harmful retrieval samples.

\subsection{Ablation: Are All Reward Functions Useful?}

In our designed methodology, there are three reward functions. To verify their indispensability, we conducted an ablation study on polluted generation by removing each reward function individually, with results shown in Table~\ref{tab:ablation_1}. 
We observed that removing the Format Reward during training causes the model's output format to deviate from specifications. 
This leads to a critical issue: while we cannot accurately assess the other two rewards for the few completion outputs violating format requirements, we could skip backpropagation for these two rewards on such completions to ignore the errors.
The Format Reward penalizes non-compliant outputs, ensuring virtually no format violations throughout training.
Without it, numerous non-compliant completions emerge, subsequently affecting the other two reward functions and causing model collapse during training.
Although removing the other two reward functions still permits stable training, the final evaluation metrics including Mean Accuracy and Accuracy Degradation demonstrate significant deterioration. 
We observed similar results in the Top-K generation experiments.
\textbf{This confirms that all three reward functions we designed are indispensable.}

\section{Conclusion}

In this work, we pose a key bottleneck in RAG systems: over-reliance on retriever precision, which often results in irrelevant or misleading contexts being fed into the generation phase.
To evaluate and mitigate this issue, we introduced \textbf{VaccineRAG}, a novel CoT-based RAG dataset, which not only serves as a benchmark for evaluating LLMs under varying positive/negative sample ratios, but also provides a scheme to enhance models’ abilities in sample discrimination through explicit CoT reasoning.
We also proposed \textbf{Partial-GRPO}, a new approach for learning multi-step, long-sequence CoT content by modeling LLM outputs as multiple components rather than a monolithic whole, enabling more nuanced preference selection and improves the model’s overall capability to handle intricate reasoning tasks.
Extensive experiments and ablation studies on VaccineRAG  highlight the importance of fine-grained reasoning and sample analysis in RAG systems, paving the way for more robust and reliable LLM applications.

\clearpage

\bibliography{ref}

\end{document}